\documentclass{article}

 \usepackage[dblblindworkshop, final]{neurips_2025}
\workshoptitle{Efficient Reasoning}

\usepackage[utf8]{inputenc} 
\usepackage[T1]{fontenc}    
\usepackage{hyperref}       
\usepackage{url}            
\usepackage{booktabs}       
\usepackage{amsfonts}       
\usepackage{nicefrac}       
\usepackage{microtype}      
\usepackage{xcolor}         
\usepackage{enumitem}
\usepackage{amsmath}
\usepackage{amssymb}
\usepackage{algorithm}
\usepackage[noend]{algpseudocode}
\algrenewcommand\algorithmicrequire{\textbf{Input:}}
\algrenewcommand\algorithmicensure{\textbf{Output:}}
\usepackage{graphicx}
\usepackage{subcaption}

\title{Beyond Static Cutoffs: One-Shot Dynamic Thresholding for Diffusion Language Models}

\author{%
  Jucheng Shen \\
  Rice University \\
  \texttt{js237@rice.edu} \\
  \And
  Yeonju Ro \\
  The University of Texas at Austin \\
  \texttt{yro@cs.utexas.edu} \\
}

\begin{document}

\maketitle

\begin{abstract}
Masked Diffusion Language Models (MDLM) are becoming competitive with their autoregressive counterparts but commonly decode with fixed steps and sequential unmasking. To accelerate decoding, recent works like Fast-dLLM enables parallel decoding via a static global confidence threshold, yet we observe strong block/step-wise confidence fluctuations and, within a dataset, near-identical confidence trajectories across inputs indicated by cosine similarity. Inspired by these two observations, we introduce \textbf{One-Shot Dynamic Thresholding (OSDT)}, which calibrates thresholds on a single sequence and applies them to subsequent inputs with negligible overhead. On GPQA, GSM8K, and HumanEval, OSDT attains superior accuracy–throughput trade-offs (\textbf{+24\%} tokens/s on GSM8K at the \textbf{best} accuracy, \textbf{+45\%} on GPQA with comparable accuracy, and \textbf{+50\%} on HumanEval with a modest accuracy gap). Beyond these results, our findings suggest broader opportunities to leverage reusable task-level confidence signatures for more general-purpose algorithmic and systems innovations in diffusion decoding. Our code is available at \url{https://github.com/jackshen-1215/osdt}.
\end{abstract}

\section{Introduction}
Masked diffusion language models (MDLM) have recently advanced rapidly, with pre-trained models such as LLaDA-8B \citep{nie2025large} and Dream-7B \citep{ye2025dream7b} scaling discrete diffusion and demonstrating strong performance across language, math, and code. Inference in these models is semi-autoregressive: the sequence is partitioned into contiguous blocks $B_1,\dots,B_T$ that are generated left-to-right (autoregressive across blocks), while within each block decoding proceeds by diffusion-style denoising steps $s=1,\dots,S$ that can unmask any subset of still-masked positions in a non-left-to-right order. At each step, a mask predictor proposes token distributions and a fixed per‑step quota of token positions is filled with topk by confidence or randomly, while the remaining positions stay masked for later steps.

To push throughput beyond fixed number of tokens per step, recent works like Fast-dLLM \citep{wu2025fastdllm} enables parallel decoding by unmasking all tokens above a \textbf{static} global confidence threshold. However, our empirical observations show that confidence is highly non-uniform: it varies across blocks and denoising steps, yet confidence trajectories are remarkably stable across inputs within the same dataset (or, the same task). The first observation calls for adaptivity over the course of generation; the second indicates that a small amount of computation can capture a task-level “confidence signature.” Together, they reveal substantial space for more efficient parallel decoding in MDLM than static schedules can offer.

We therefore study \textbf{dynamic thresholding for parallel decoding in masked diffusion models}. Formally, given a remasking decoder and per-step confidence scores, the goal is to design an inference-time policy $\pi$ that selects thresholds $\{\tau_{b}\}$ or $\{\tau_{b,s}\}$ to decide which masked tokens to unmask, maximizing throughput while maintaining task accuracy (or, equivalently, improving the accuracy–throughput Pareto frontier). The policy must be \textbf{task-aware, training-free, and add negligible overhead}.

Our contributions can be summarized as follows:
\begin{itemize}[topsep=0pt]
    \item We empirically establish that within a dataset the block/step-wise confidence vectors are highly correlated across inputs (cosine similarity near 1), revealing a stable task-level signature. 
    \item We introduce \textbf{One-Shot Dynamic Thresholding (OSDT)}, a two-phase decoder that calibrates thresholds on a single sequence and reuses them for subsequent inputs at either block or step-block granularity. 
    \item We provide a comprehensive evaluation on GPQA, GSM8K, and HumanEval, showing that OSDT consistently improves tokens/s at comparable accuracy and often sets a better Pareto frontier than static thresholds as in Fast-dLLM.
\end{itemize}

\section{Observation}

\begin{figure}[h]
  \centering
  \includegraphics[width=1.0\linewidth]{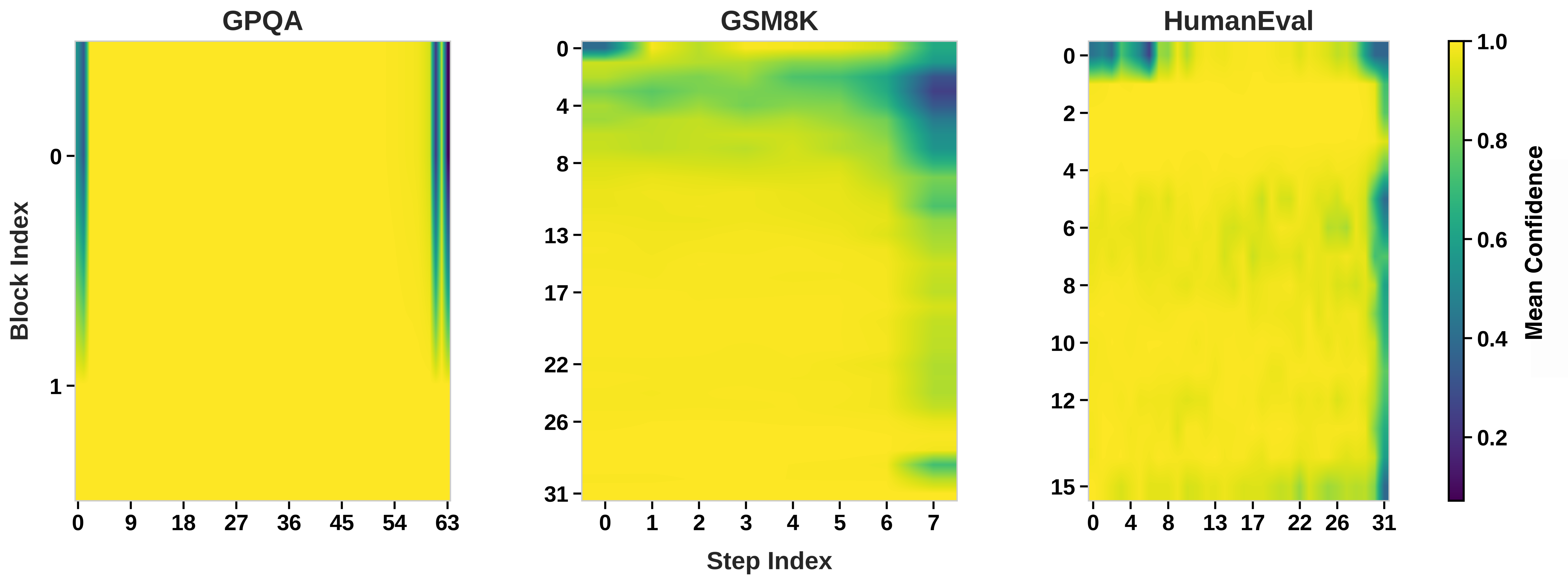}
  \caption{\textbf{Step-block mean token confidence.} 
  Across GPQA, GSM8K, and HumanEval, confidence starts low, peaks mid-process, and drops near the final steps. These structured U-shaped dynamics highlight the limits of static thresholding.}
  \label{fig:conf}
\end{figure}

Prior work~\cite{wu2025fastdllm} assumes confidence follows a generalizable pattern and applies a static threshold across tasks. In practice, as shown in Figure~\ref{fig:conf}, benchmarks as different as GPQA (expert-level Q\&A), GSM8K (grade-school math), and HumanEval (code generation) exhibit distinct signatures. This task-dependent variability undermines any “one-size-fits-all” thresholding strategy, highlighting the need for adaptation at inference time.

A natural question is whether these dynamics remain stable within a dataset. If confidence patterns are consistent across inputs, the profile of one sequence could proxy for the rest. To test this, we measured cosine similarity between average step-blockwise confidence vectors for all input pairs.

\begin{figure}[h]
  \centering
  \includegraphics[width=1.0\linewidth]{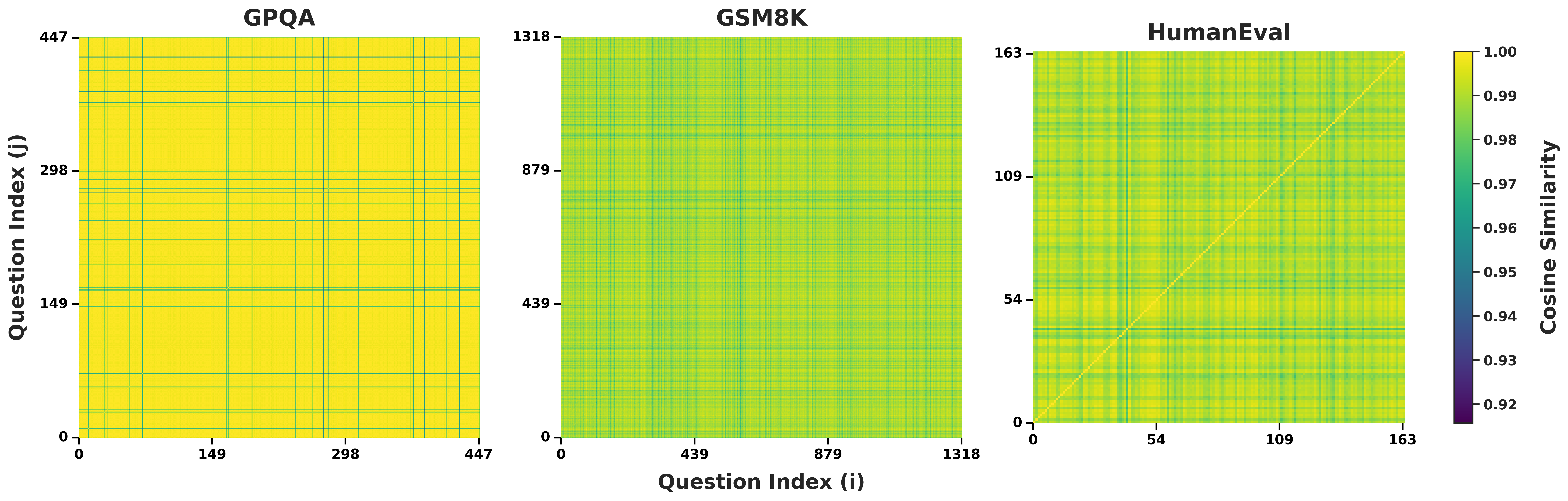}
  \caption{\textbf{Pairwise cosine similarity of step-block mean token confidence.} 
  Confidence trajectories are nearly identical across inputs of the same dataset, suggesting a single calibration run can generalize to the entire benchmark.}
  \label{fig:cos_sim_stepblock}
\end{figure}

Results in Figure~\ref{fig:cos_sim_stepblock} show striking consistency: cosine values are near 1.0 across GPQA, GSM8K, and HumanEval, producing nearly uniform bright heatmaps. This indicates that while confidence levels differ across tasks, the relative trajectories are highly stable within a dataset—a task-level rather than instance-level property.

This stability implies strong predictive power: the behavior on one sequence forecasts that of others in the same dataset. It provides the basis for adaptive thresholding methods that calibrate once and generalize broadly, enabling more efficient inference without sacrificing accuracy.

\section{Dynamic Threshold}

These findings expose the limits of static cutoffs: they fail to adapt to structured, task-dependent confidence dynamics, yet within-dataset patterns are highly consistent. This motivates a method that is both dynamic and sample-efficient—able to adapt thresholds to a dataset’s confidence profile with minimal overhead.

We propose \textbf{One-Shot Dynamic Thresholding (OSDT)}, a two-phase approach:

\begin{itemize}[topsep=0pt,leftmargin=*]
    \item \textbf{Phase 1 (Calibration):} The first sequence is decoded with standard static-thresholding as in Fast-dLLM, and block- or step-block-wise confidence vectors are collected.
    \item \textbf{Phase 2 (Dynamic Inference):} Subsequent sequences use thresholds derived from the calibration profile, with safeguards such as a cap $\kappa$ and slack ratio $\epsilon$ to balance efficiency and robustness.
\end{itemize}

This mechanism dynamically aligns decoding decisions to the task’s confidence landscape. Unlike static thresholding, OSDT tailors thresholds to dataset-level dynamics while incurring negligible extra cost. Algorithm~\ref{alg:osdt} (see Appendix) outlines the procedure. Its reliance on a single calibration run makes OSDT practical, efficient, and well-suited to real-world scenarios where both accuracy and latency matter.

\section{Evaluation}

We evaluate \textbf{One-Shot Dynamic Thresholding (OSDT)} on three benchmarks: GPQA (expert-level Q\&A), GSM8K (grade-school math), and HumanEval (code generation). OSDT calibrates a confidence profile on the first sequence of a dataset and applies dynamically adjusted thresholds to subsequent inputs. We study both its hyperparameter behavior and its performance against Fast-dLLM \citep{wu2025fastdllm}. All experiments use batch size 1 on a single NVIDIA H100 GPU.

\subsection{Hyperparameters}

OSDT exposes four hyperparameters that trade accuracy for throughput:
\begin{itemize}[topsep=0pt,leftmargin=*]
    \item \textbf{Dynamic Mode ($M$):} Threshold per block (\emph{block}) or per denoising step within each specific block (\emph{step-block}).
    \item \textbf{Threshold Metric ($\mu$):} Statistic over calibration confidences (mean, Q1, median, Q3, min-whisker).
    \item \textbf{Threshold Cap ($\kappa$):} Upper bound to avoid overly strict thresholds.
    \item \textbf{Slack Ratio ($\epsilon$):} Downscales thresholds to increase parallelism.
\end{itemize}

We perform a grid search over {\small $\mu,\ \kappa\!\in\!\{0.75,0.8,0.85,0.9,0.95\},\ \epsilon\!\in\!\{0.01,0.05,0.1,0.15,0.2\}$}. Results (see Appendix for more details) show task-dependent optima: GPQA benefits from fine-grained \emph{step-block} thresholds, whereas GSM8K and HumanEval prefer the simpler \emph{block} mode. We use the following configurations in all following comparisons:
\begin{itemize}[topsep=0pt,leftmargin=*]
    \item \textbf{GPQA:} \emph{step-block}, $q2$, $\kappa=0.75$, $\epsilon=0.20$.
    \item \textbf{GSM8K:} \emph{block}, $q1$, $\kappa=0.75$, $\epsilon=0.20$.
    \item \textbf{HumanEval:} \emph{block}, $q1$, $\kappa=0.80$, $\epsilon=0.10$.
\end{itemize}

\subsection{Comparative Results}

Table~\ref{tab:main_comparison} compares OSDT to Fast-dLLM’s fixed-threshold ($\tau=0.9$) and best factor-based settings. OSDT consistently achieves a better accuracy–throughput trade-off across diverse reasoning and code generation benchmarks:

\begin{itemize}[topsep=0pt,leftmargin=*]
    \item \textbf{GSM8K:} OSDT achieves the highest accuracy (76.0\%) while running 24\% faster than the fixed-threshold baseline, indicating that dynamic thresholding effectively balances precision and decoding speed for step-by-step numerical reasoning.
    \item \textbf{GPQA:} OSDT maintains comparable accuracy (29.2\% vs.\ 29.9\%) with 45\% higher throughput, suggesting that OSDT adapts well even in multi-hop question answering tasks with varying reasoning depth.
    \item \textbf{HumanEval:} OSDT delivers similar pass rate (40.9\% vs.\ 43.3\%) yet achieves 50\% faster throughput, demonstrating that dynamic control does not compromise code-generation reliability.
\end{itemize}

Overall, OSDT achieves more favorable Pareto points, accelerating inference without significant quality degradation. The results highlight that adaptive thresholding improves computational efficiency by selectively reducing redundant forward passes—particularly when model confidence stabilizes—yielding consistent gains across both reasoning and code generation domains.

\begin{table}[h!]
\centering
\caption{\textbf{Comparative results.} Best values in \textbf{bold}. Throughput in tokens/s.}
\label{tab:main_comparison}
\renewcommand{\arraystretch}{1.05}
\setlength{\tabcolsep}{5.2pt}
\small
\begin{tabular}{lcccccc}
\toprule
\textbf{Benchmark} &
\multicolumn{2}{c}{\textbf{OSDT (Ours)}} &
\multicolumn{2}{c}{\textbf{Fast-dLLM (Fixed)}} &
\multicolumn{2}{c}{\textbf{Fast-dLLM (Factor)}} \\
\cmidrule(lr){2-3} \cmidrule(lr){4-5} \cmidrule(lr){6-7}
& \textbf{Acc. (\%)} & \textbf{Thru.} & 
  Acc. (\%) & Thru. & 
  Acc. (\%) & Thru. \\
\midrule
GPQA      & 29.24 & \textbf{63.27} & 28.12 & 42.69 & \textbf{29.91} & 43.58 \\
GSM8K     & \textbf{76.00} & \textbf{230.75} & 74.75 & 172.74 & 75.00 & 186.63 \\
HumanEval & 40.85 & \textbf{172.25} & 39.63 & 152.51 & \textbf{43.29} & 114.71 \\
\bottomrule
\end{tabular}
\end{table}

\section{Related Work}


Recent advances in masked diffusion language models (DLMs) demonstrate their potential as an alternative to autoregressive generation. \textbf{LLaDA} \citep{nie2025large} scales discrete diffusion to 8B parameters with block-wise decoding and low-confidence remasking, while \textbf{Dream} \citep{ye2025dream7b} improves training via AR initialization and context-adaptive token-level noise rescheduling. Both rely on fixed-step schedules that unmask tokens one by one, limiting efficiency.

\textbf{Fast-dLLM} \citep{wu2025fastdllm} addresses this by formally proving when greedy parallel decoding with product-of-marginals is equivalent to sequential decoding in high-confidence regimes, enabling safe parallel generation. It then proposes a \emph{confidence-aware parallel decoding} rule that unmasks all tokens above a global static cutoff, and introduces prefix and dual (prefix+suffix) KV-Cache designs to improve throughput. However, its thresholding remains task-agnostic and static.

In contrast, our work focuses on \emph{dynamic, task-aware thresholding}. Instead of relying on fixed schedules or static cutoffs, we calibrate thresholds from a single input and adapt them to dataset-level confidence patterns, yielding more efficient and accurate decoding.

\section{Conclusion}

In this paper, we identify a simple but overlooked property of MDLMs: confidence evolves in structured ways over time yet is strikingly consistent across inputs within the same task. Building on this, we introduce \textbf{One-Shot Dynamic Thresholding (OSDT)}, a training-free, dataset-aware decoding scheme that calibrates thresholds on a single sequence and applies them with negligible overhead; 
On LLaDA-8B across GPQA, GSM8K, and HumanEval, OSDT improves the accuracy–throughput frontier, e.g., +24\% tokens/s at the best accuracy on GSM8K, +45\% throughput on GPQA at comparable accuracy, and +50\% on HumanEval with a modest accuracy gap. 



\small
\bibliographystyle{unsrt}
\bibliography{ref}

\appendix

\section{Technical Appendices and Supplementary Material}

\subsection{OSDT Algorithm}

\begin{algorithm}[h]
\caption{One-Shot Dynamic Thresholding (OSDT)}
\label{alg:osdt}
\begin{algorithmic}[1]
\Require Prompts $Q=\{q_1,\dots,q_n\}$; mode $M\in\{\texttt{block},\texttt{step-block}\}$; metric $\mu$; cap $\kappa$; slack $\epsilon$
\Ensure Answers $A=\{a_1,\dots,a_n\}$
\State $A\gets\varnothing$;\quad $\mathcal{T}\gets\varnothing$
\Statex \textit{Notation: } $\mathcal{B}(b)$ denotes the indices in block~$b$.
\For{$i\gets 1$ \textbf{to} $n$}
  \If{$i=1$} \Comment{one-shot calibration}
    \State $(a_1,\textit{conf})\gets\textsc{StandardGenerate}(q_1)$
    \State $\mathcal{T}\gets\textsc{Calibrate}(\textit{conf}, M, \mu)$
    \State \Call{Append}{$A, a_1$}
  \Else
    \State $x\gets\textsc{InitSequence}(q_i)$
    \For{$b\gets 1$ \textbf{to} $\textit{num\_blocks}$}
      \State $\textit{step}\gets 0$
      \While{\textsc{Masked}$(x,b)$}
        \State $\textit{conf}\gets\textsc{Confidence}(\textsc{Predict}(x),\mu)$
        \If{$M=\texttt{step-block}$}
          \State $\tau\gets\mathcal{T}[b][\textit{step}]$ \Comment{step-block-level}
        \Else
          \State $\tau\gets\mathcal{T}[b]$ \Comment{block-level}
        \EndIf
        \State $\tau\gets\min(\tau,\kappa)$;\quad $\tau_{\mathrm{eff}}\gets\tau(1-\epsilon)$
        \State $S\gets\{\, j\in\mathcal{B}(b)\;:\; \textit{conf}[j]>\tau_{\mathrm{eff}} \,\}$ \Comment{indices in block~$b$ above threshold}
        \If{\Call{IsEmpty}{$S$}}
          \State $\textit{idx}\gets\textsc{IndexOfMax}(\textit{conf},\,\text{block }b)$
          \State $S\gets\{\textit{idx}\}$ \Comment{fallback: unmask most confident index in block~$b$}
        \EndIf
        \State \textsc{UnmaskAndUpdate}$(x,S)$;\quad $\textit{step}\gets\textit{step}+1$
      \EndWhile
    \EndFor
    \State $a_i\gets x$;\quad \Call{Append}{$A, a_i$}
  \EndIf
\EndFor
\State \Return $A$
\end{algorithmic}
\end{algorithm}

\subsection{Full Hyperparameter Sweep Results}

We provide detailed accuracy–throughput trade-offs for OSDT across GPQA, GSM8K, and HumanEval. Each plot visualizes all combinations of dynamic mode ($M$), threshold metric ($\mu$), threshold cap ($\kappa$), and slack ratio ($\epsilon$). Marker shape denotes $\mu$, marker size denotes $\kappa$, color indicates $\epsilon$, and line style distinguishes \emph{step-block} (solid) from \emph{block} (dashed).

\begin{figure}[h]
  \centering
  \includegraphics[width=1.0\linewidth]{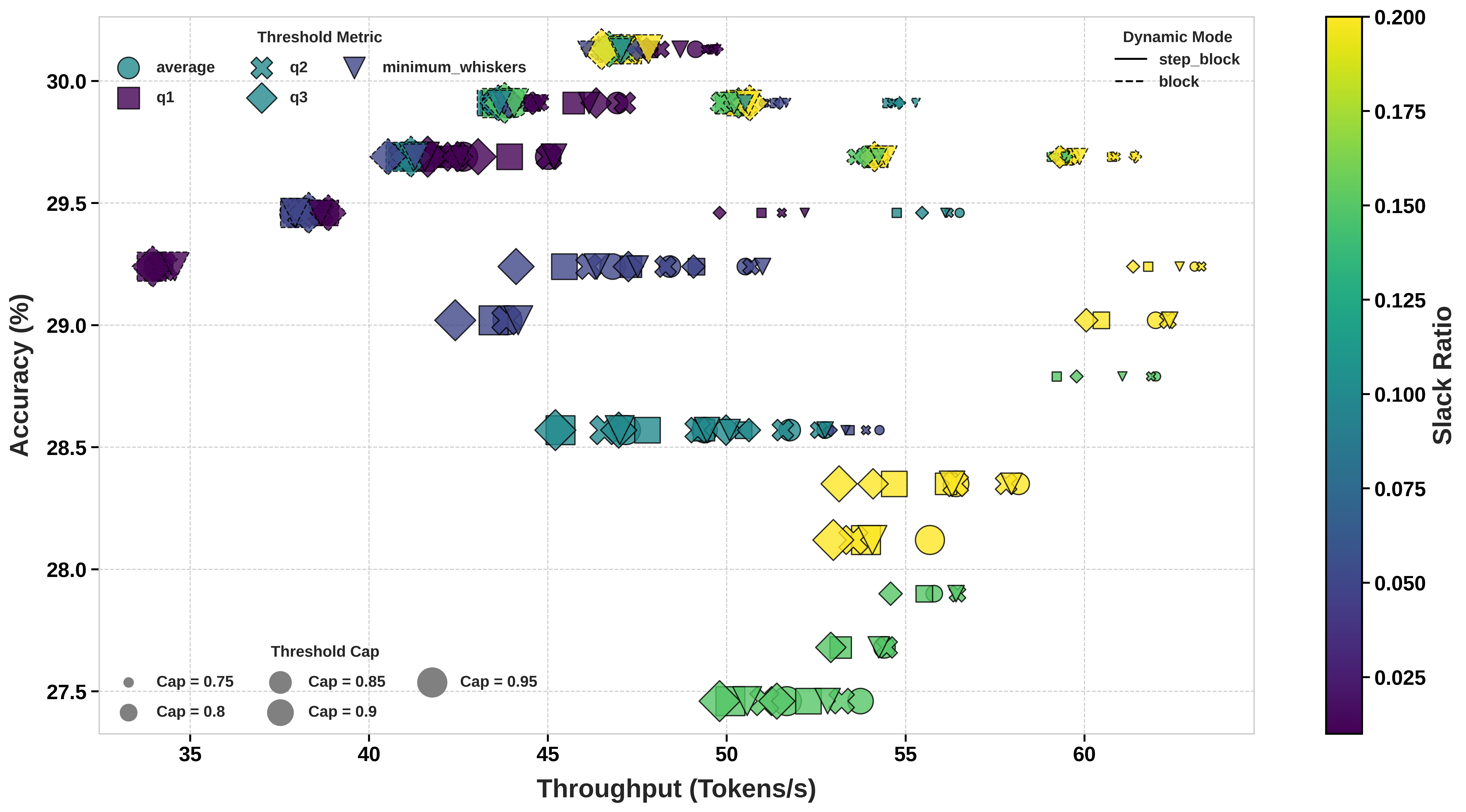}
  \caption{\textbf{GPQA hyperparameter sweep.} 
  Accuracy peaks near 30\% but varies only slightly across settings, while throughput is strongly influenced by $\epsilon$ and $\kappa$. Step-block mode provides finer adaptation and better trade-offs in high-accuracy regions.}
  \label{fig:gpqa_sweep_appendix}
\end{figure}

\begin{figure}[h]
  \centering
  \includegraphics[width=1.0\linewidth]{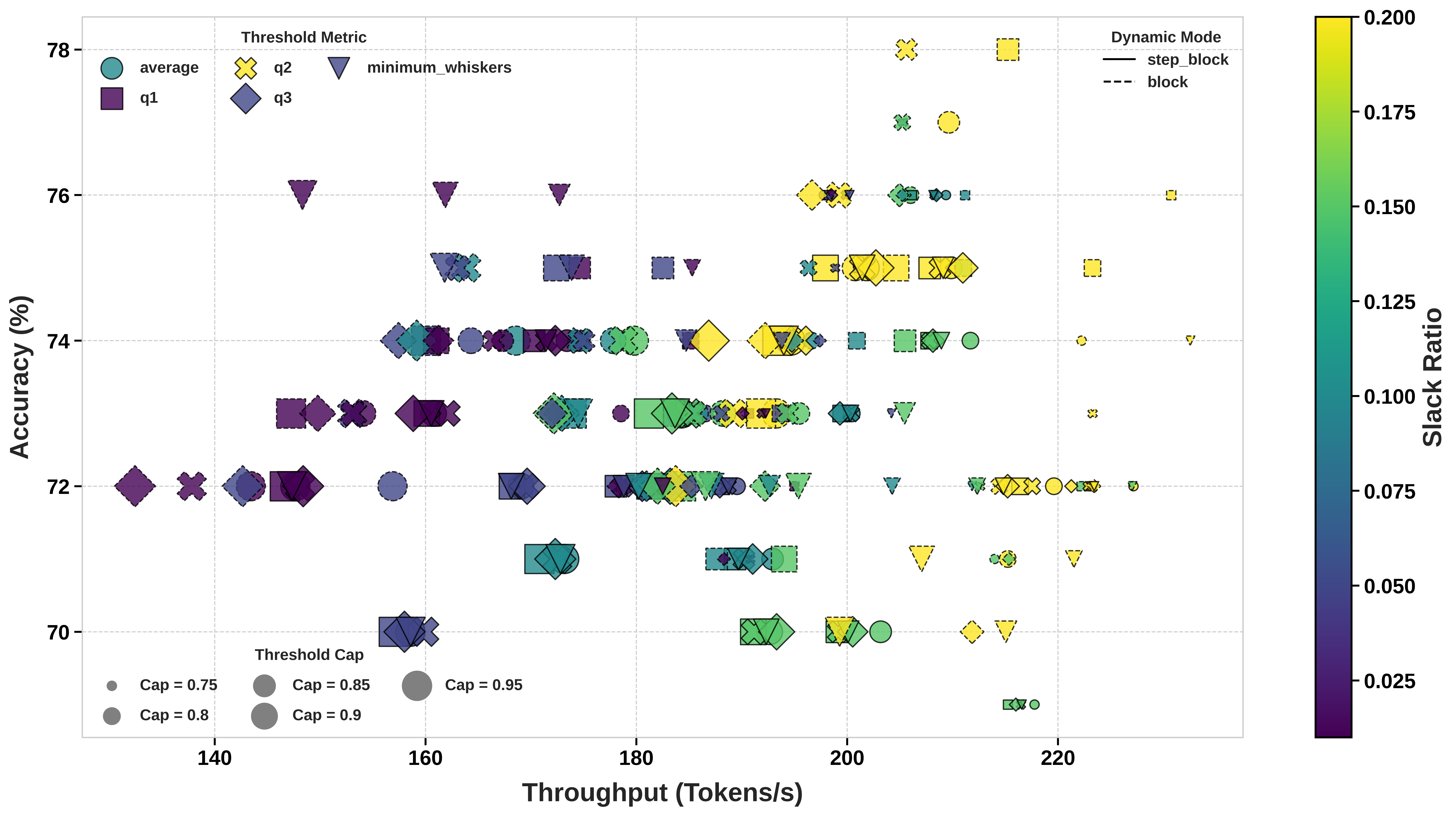}
  \caption{\textbf{GSM8K hyperparameter sweep.} 
  Structured reasoning tasks benefit most from block-level thresholds, which achieve higher accuracy (up to 76\%) while maintaining strong throughput. Step-block offers little advantage here, confirming block mode suffices for GSM8K.}
  \label{fig:gsm8k_sweep_appendix}
\end{figure}

\begin{figure}[h]
  \centering
  \includegraphics[width=1.0\linewidth]{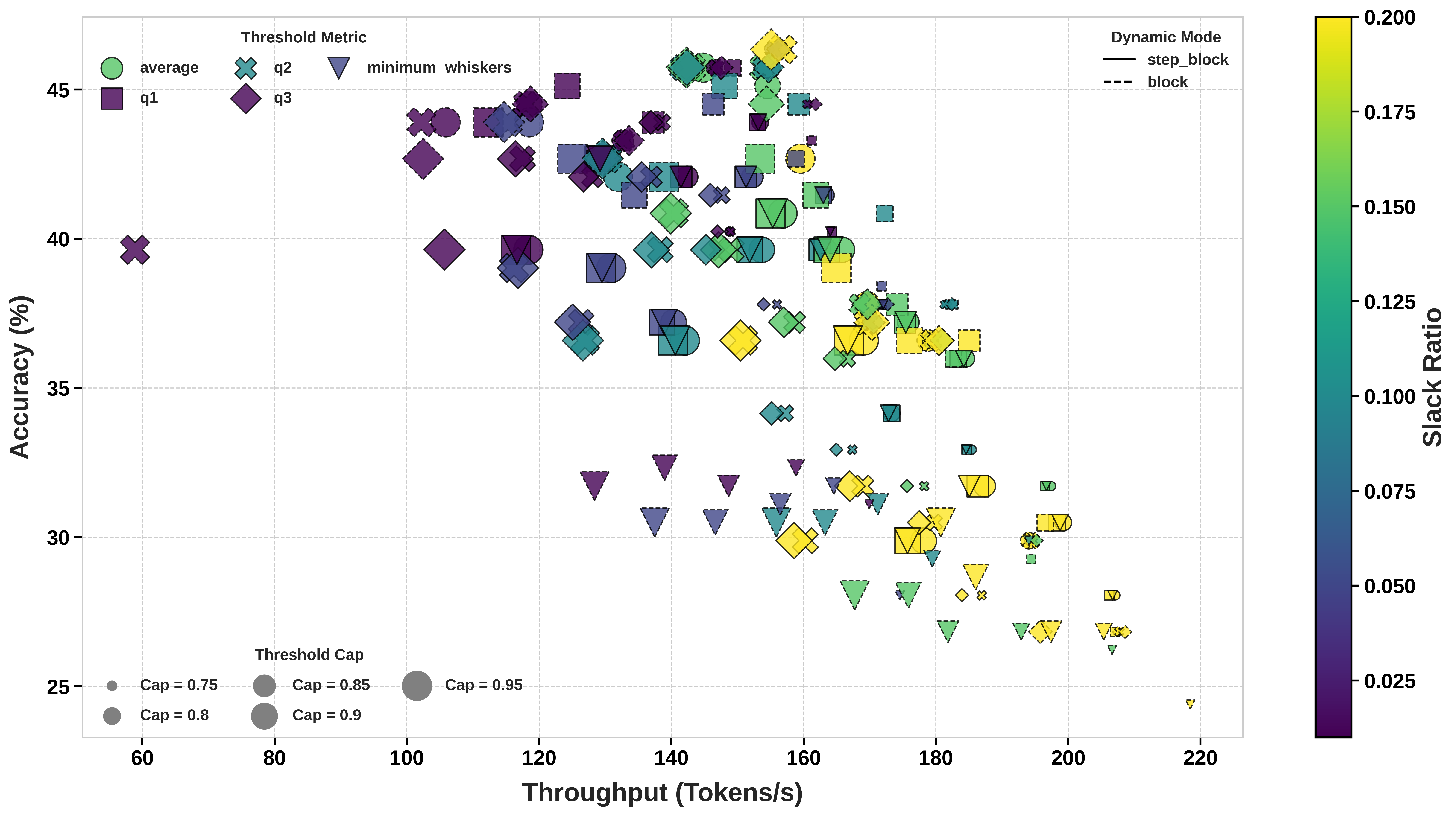}
  \caption{\textbf{HumanEval hyperparameter sweep.} 
  Code generation shows a sharper accuracy–throughput trade-off: aggressive settings yield large speedups but accuracy drops quickly. Block-level thresholds dominate the Pareto frontier, offering simpler yet more efficient schedules.}
  \label{fig:humaneval_sweep_appendix}
\end{figure}

\end{document}